\documentclass{article}

\usepackage{PRIMEarxiv}

\usepackage{lineno, hyperref}
\RequirePackage{fix-cm}

\usepackage{graphicx}
\usepackage[misc]{ifsym}
\usepackage{blindtext}

\usepackage[draft]{todonotes}
\usepackage{amsmath}
\usepackage{bm}
\usepackage{amssymb}
\usepackage{caption}
\usepackage{subcaption}
\usepackage{hyperref}
\usepackage{graphicx}
\usepackage{pgfplots}
\usepackage{verbatim}
\usepackage{graphics}
\usepackage{algorithm}
\usepackage{algorithmic}
\usepackage{placeins}
\usepackage{float}
\usepackage{paralist}

\usepackage[utf8]{inputenc} 
\usepackage[T1]{fontenc}    
\usepackage{booktabs}       
\usepackage{amsfonts}       
\usepackage{nicefrac}       
\usepackage{microtype}      

\DeclareMathOperator*{\argmax}{\arg\max}

\title{Contextual Decision Trees
}

\author{
  Tommaso Aldinucci, \ Enrico Civitelli, \ Leonardo Di Gangi, \ Alessandro Sestini \\
  Department of Information Engineering \\
  Universit\`{a} degli Studi di Firenze \\
  Via di Santa Marta 3, 50139, Florence, \\
  \texttt{\{tommaso.aldinucci,\ enrico.civitelli,\ leonardo.digangi,\ alessandro.sestini\}@unifi.it} \\
}

\begin{document}
\maketitle

\begin{abstract}
Focusing on Random Forests, we propose a multi-armed contextual bandit recommendation framework for feature-based selection of a single shallow tree of the learned ensemble. The trained system, which works on top of the Random Forest, dynamically identifies a base predictor that is responsible for providing the final output. In this way, we obtain local interpretations by observing the rules of the recommended tree. The carried out experiments reveal that our dynamic method is superior to an independent fitted CART decision tree and comparable to the whole black-box Random Forest in terms of predictive performances.
\end{abstract}

\keywords{Random Forest \and Interpretability \and Reinforcement Learning}

\section{Introduction}

Supervised learning algorithms are tuned to optimize their predictive accuracy often at the cost of compromising the understanding of the model. However, in some domains involving critical decision making such as health care, law or banking, the need for insight into the reasons of a specific prediction is pronounced \cite{survey_explainability}. In other words, supervised learning models need to be interpretable, in the sense that they must be able to answer the question ``Why does this particular input lead to that particular output?'' \cite{gilpin2018}. In order to address this latter need, the use of interpretable models is recommended whenever possible \cite{Rudin2019Stop}. Nevertheless, especially when the focus is to obtain high predictive power and the employment of black box models is unavoidable as in the case of structured data, approaches that aim to locally explain predictions are spreading out \cite{ribeiro-etal-2016-trust}, \cite{KENNY2021107530}, \cite{BLANCOJUSTICIA2020105532}. In a sense, this allows to obtain a-posteriori explanations of the predictions, while the direct employment of interpretable models as linear or decision trees models enables for itself the understanding of predictions, i.e. the interpretability is intrinsic. This entire set of processes and methods constitutes the so called AI explainability research area \cite{molnar2020interpretable}. 

On the other hand, especially in time series forecasting \cite{lemke2010meta}, \cite{talagala2018meta}, \cite{montero2020fforma}, \cite{digangi} as well as in a wide variety of situations like Active Model Selection \cite{madani2012active} and Active Learning \cite{karimi2021online}, the problem of automatically selecting an effective predictive algorithm is relevant. This represents an instance of The Algorithm Selection Problem, introduced and formalized in \cite{rice1976algorithm}. In the more recent machine learning community, the methods that seek to exploit knowledge about learning algorithm performance to improve the performance or selection of learning algorithms itself are known as meta-learning methods \cite{smith2009cross}.

Against this background, in this paper we focus on designing a learning algorithm which, given an input instance, recommends a single shallow tree of a trained Random Forest. The proposed system aims to learn a mapping which links the input feature space to the space of grown decision trees in order to identify an interpretable and effective predictor for that input sample. We clarify that the proposed recommendation system is itself a black-box model but its recommendation leads to an interpretable model. More specifically, although our method leads to post-hoc local interpretability, the recommended tree does not act as a surrogate explainer of the whole black-box Random Forest. Our intuition is that the presence of such a black-box layer, which guarantees a dynamic and data-driven selection of decision trees, can approximate the predictive performances of the Random Forest. Due to this context-based nature, we will refer to our approach as Contextual Classification Tree (CCT) or Contextual Regression Tree (CRT) in the remainder of this manuscript for, respectively, classification and regression tasks.

More in detail, the problem is addressed within a Contextual Multi-Armed Bandit framework, which represents a subclass of Markov Decision Processes (MDP) with an unitary length of each episode \cite{gampa2019banditrank}. The usage of Multi-Armed Bandit models for recommendations is well known in the literature and it is employed in fields such as advertisements, personalized news articles, healthcare and finance \cite{ZHANG2022107922} \cite{BanditNews} \cite{ContextualCancer}, \cite{Shen2015PortfolioCW}.

In our case, the action taken by the agent consists in the choice of one tree of the Random Forest on the basis of the input features which act like the observed context. This choice is determined by a parametrized policy which encodes a probability distribution over the trees of the Random Forest. After each choice, the agent receives a reward that depends both on the context and the selected action. Considering that we focus on prediction, the adopted reward is related to the predictive ability of the chosen tree: high reward if the selected one obtains a low prediction error and vice-versa.

Parameters of policy are learnt, using policy gradient Reinforcement Learning (RL) methods \cite{gampa2019banditrank}, in order to maximize the cumulative rewards over a certain period. Gradient methods for bandits are well motivated in \cite{sutton1998rli} and can be applied also in the contextual case \cite{policyRecommendation}. Moreover, policy gradient methods have good convergence property for finite states and finite actions MDPs \cite{convergencePolicyGradient}.

In the current formulation, our policy is a lightweight neural network. The choice of a RL approach over a supervised one is motivated noting that the labelling process, besides being metric dependent, would finish to \textit{a-priori} exclude some of the Random Forest predictors.  

On the other hand, RL ensures a wider and more robust exploration phase by means of its trial-error learning approach and the \textit{exploration}/\textit{exploitation} trade-off.

Our contribution is related to the design of such a system. The proposed method is very general and can be applied in many contexts where personalized recommendations are needed (finance, healthcare, advertisements etc...) \cite{bouneffouf2019survey}. Moreover, although our method is based on a Random Forest, it can be extended to any ensemble of predictors, not necessarily interpretable predictors. However, we reckon that our approach is relevant for the purpose of obtaining improved interpretable models.     


\vspace{0.5em}
The rest of the paper is organized as follows: 
\begin{itemize}
    \item In Section \ref{sec:related_works} we report a detailed summary of the existing methodologies that have similarities with the introduced method.   
    
    \item In Section \ref{sec:reinforce} we provide a brief introduction to the basic ideas and intuitions covering the area of RL which is related to our approach.
    
    \item In Section \ref{sec:proposed_method} we describe our method in detail. 
    
    \item In Section \ref{sec:numerical_experiments} we report experimental results on real-world datasets for both binary classification and regression tasks. 
    
    \item In Section \ref{sec:conclusion} we summarize our contribution and we suggest some possible future developments.
\end{itemize}

\section{Related Works} \label{sec:related_works}
Our approach relates to several bodies of literature.

As mentioned in the introduction, meta-learning methods, which automate the model selection task, arise especially in the area of time series forecasting. In fact, forecasting problems often possess a small and finite history of points to consistently select a predictive model, and meta-learning methods are of considerable support in automating the model selection phase and improving the predictive performances. In particular, both black-box \cite{talagala2018meta}, \cite{montero2020fforma}, \cite{digangi} and interpretable models \cite{lemke2010meta} are used as meta-learners. Unlike our own approach, these meta-learners are trained by supervised learning techniques and are mainly designed to obtain forecasts as a convex combination of a pool of forecasting methods. In common with our approach is the fact that these meta-learners perform model selection at a feature level: each time series is mapped into a set of representative and meaningful features.                 

Algorithm recommendation methods \cite{brazdil2008metalearning} are strongly related to our approach. However in this case, the algorithms are recommended at a dataset level rather than at a feature level.

Similarly to our case, a set of pre-trained classifiers is also available in Active Model Selection \cite{madani2012active} and Active Learning \cite{karimi2021online}. However, in both cases the aim is to identify the best model for new samples from the set of fitted models within budgetary constraints.

Furthermore, our methodology is linked to all the methods \cite{vidal2020born}, \cite{sagi2020explainable} whose aim is to obtain a single decision tree from a Random Forest. In particular, a set of rule conjunctions that represent the original forest is first created in \cite{sagi2020explainable}; the conjunctions are then hierarchically organized to form a new interpretable decision tree. In the context of explaining Random Forest, in \cite{vidal2020born} a method is proposed to construct a single predictor tree from the entire forest. Specifically, they employed a dynamic-programming based algorithm that exploits pruning and bounding rules. By design, the obtained tree has the same predictive behaviour of the associated Random Forest although it has obviously more leaves.

In the context of deep learning the conditional computation approach \cite{bengio2015conditional} has similarities with our method. In fact, in this latter case, in an input-dependent way, similar to our approach, RL is used to activate only some of the units in a network for faster inference.    

Finally, there are also RL approaches for constructing decision trees. In particular, in \cite{xiong2017learning} a strategy, based on long-term rewards, is proposed to search for proper ways to split the feature space. They showed that the obtained tree outperforms the standard greedy building strategy of CART in several datasets.

\section{Reinforcement Learning: Preliminaries and Notations} \label{sec:reinforce}
Reinforcement Learning (RL) aims to train autonomous agents in order to learn behavior through trial-error interactions with a dynamic environment. The agent choose an action at each time step, which changes the state of the environment in an unknown way, and receives feedback based on the consequence of the action.

RL can be formally defined as a Markov Decision Process (MDP) \cite{arulkumaran2017brief} consisting of:
\begin{itemize}
    \item a set of states $\mathcal{S}$;
    
    \item a set of actions $\mathcal{A}$;
    
    \item transition dynamics $d(s_{t+1} | s_t, a_t)$ mapping a state, action 
    couple at time $t$ into a distribution of possible state at time $t + 1$;
    
    \item a Reward Function $$r_{t+1} = r(s_t, a_t, s_{t+1}): \mathcal{S} \times \mathcal{A} \times \mathcal{S} \rightarrow \mathbb{R};$$
    
    \item a discount factor $\gamma \in [0, 1)$, where lower values place more emphasis on immediate rewards.
\end{itemize}

In addition to this, RL models are based on the use of a policy $\pi$:
$$\pi: \mathcal{S} \rightarrow p(\mathcal{A}=a|\mathcal{S})$$
i.e. a mapping from states to a probability distribution over the set $\mathcal{A}$ of actions, which determines the behavior of the agent in the choice of adequate actions. In the reminder we consider episodic MDP, where the agent interacts with the environment  repeatedly in episodes of fixed length $T$. Each episode is characterized by a a sequence of states and actions $\tau = (s_t, a_t)_{t=0}^{T-1}$, usually called \textit{trajectory}. The accumulated reward $R(\tau)$ for a trajectory is given as  

$$R(\tau) = \sum_{t=0}^{T-1} \gamma^{t} r_{t+1}.$$

We stress the fact that the policy affects the probability of a given trajectory, i.e. $\tau = \tau_{\pi}$. 

Policy search methods, which represent one of the main approaches to solve RL problems \cite{arulkumaran2017brief}, are conceived to find the best policy $\pi^*$ that maximizes the expected return over all possible \textit{trajectories}:
$$\pi^* = \argmax_{\pi} \mathbb{E}_{\tau \sim \pi}[R(\tau)].$$
Since the expectation works w.r.t. the distribution of trajectories in the environment, the optimal policy $\pi^*$ does not depend on the states. 

Policy search methods use parametrized policies $\pi_{\theta}$ and in deep RL \cite{arulkumaran2017brief}, deep neural networks are employed to approximate these policies. The network outputs a probability distribution over the set of action $A$ and the action with the highest probability will define the most probable move that the policy should do in order to get the highest final reward. In this setting, policy parameters $\theta$ are updated using a gradient-based approach in order to maximize the expected return:      



$$J_\theta = \mathbb{E}_{\tau \sim p_{\theta}(\tau)}[R(\tau)] .$$
Thus, during the learning phase, at each optimization step $k$ the parameters are updated using the rule:

$$\theta_{k+1} = \theta_k + \alpha_k \nabla_\theta J_\theta.$$

Where $\alpha_k$ is the learning rate at step $k$. Gradient is computed by means of the REINFORCE algorithm \cite{williams1992simple}, which belongs to the class of Likelihood-ratio methods \cite{deisenroth2013survey}. These methods make use of the so called ``likelihood-ratio'' trick\footnote{The gradient w.r.t. $\theta$ of an expectation over a function $f(\cdot)$ of a random variable $X \sim p(x| \theta)$ is: $$ \nabla_{\theta} \mathbb{E}_X \big[f(X; \theta) \big] = \mathbb{E}_X \big[f(X; \theta) \nabla_{\theta} \log p(x|\theta) \big]$$}.

For the sake of completeness, we clarify that the loss function $\mathcal{L}_{\theta}$, used to train the Policy $\pi_\theta$, is 
$$\mathcal{L}(\theta) = -\frac{1}{N}\sum_{i=1}^{N} \sum_{t=0}^{T-1}\log\pi_\theta(a_t | s_t) \cdot R(\tau^{(i)}),$$
where $N$ is the number of episodes that define a training batch.

\section{Proposed Method} \label{sec:proposed_method}

\begin{figure}[!ht]
    \centering
    \includegraphics[width=0.35\columnwidth]{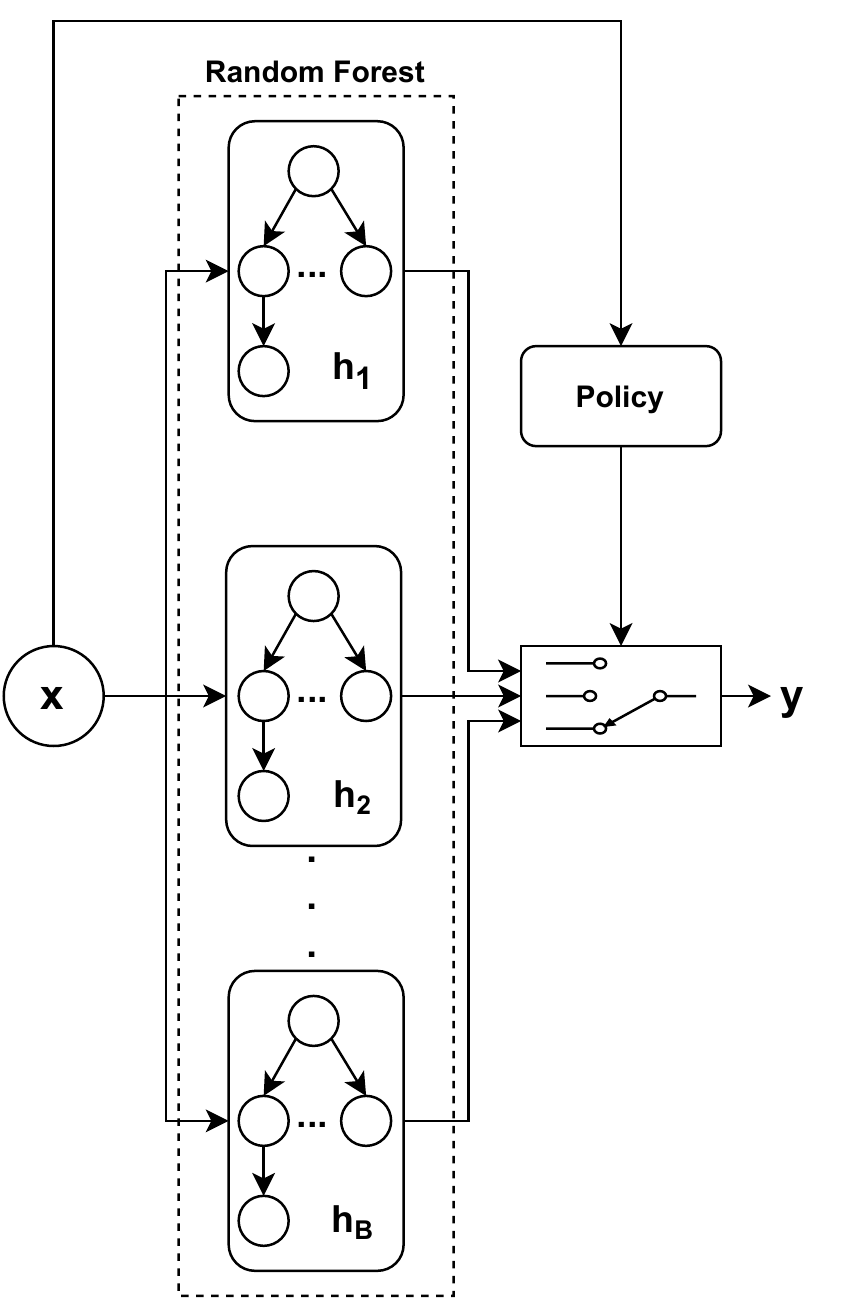}
    \caption{A high level view of our system: a Deep Neural Network (DNN) based policy selects the tree of a trained Random Forest that will make the final prediction. Each data point $x$ is observed as context by the agent and the policy aims to recommend the best tree of the Random Forest in terms of predictive performances.}
    \label{fig:blocks_ddt_drt_1}
\end{figure}

Our method, as summarized in Fig. \ref{fig:blocks_ddt_drt_1}, exploits the structure of Random Forests to provide a general locally-interpretable and feature-based recommendation system.
Given a Random Forest $RF = \{h_{1}, h_{2}, \dots, h_{B}\}$ with $B$ trees, the aim of our approach consists of training an autonomous agent to learn a policy of tree
recommendation which, given an input point $x \in \mathcal{X}$ as context and a state $s \in \mathcal{S}$ returns a probability distribution over the indexes of the trained trees of the Forest $\mathcal{T} = \{1, 2, \dots, B\}$: 
\begin{equation*} \label{eq: mapping}
    \pi(\cdot): \mathcal{X} \times \mathcal{S} \to p(\mathcal{A}=a|\mathcal{S}, \mathcal{X})
\end{equation*}
Therefore, our action space $\mathcal{A}$ is the discrete index set $\mathcal{T}$ of all the possible $B$ trees in the Random Forest.

As mentioned above, our scenario can be viewed as a contextual multi-armed bandit problem. Contextual bandit is a variant of the bandit problem, where at each episode $i$ the agent conditions its action $a_{i}$ on the context $x_i$ of the environment and observes the reward $r(a_i)$ for the chosen action. 
Moreover, it is important to note that the action affects only the immediate reward and for this reason our formulation is one-state and one-step episodic, meaning that our episode is made by only one step and $a_i$ does not condition the next state which will be always the same.
Hence, for this particular set up, the following hold:
\begin{align*}
& T = 1, \quad s_t = s, \quad a_t = a_{x_{i}}= a_i, \\
& R(s_t, a_t) = R(a_{x_i}) = r(a_i), \\
& \pi(\cdot): \mathcal{X} \to p(\mathcal{A}=a|\mathcal{X})
\end{align*}
Note that in our case, since $T = 1$ and there is only one state, the reward depends only on the action taken given the context. Thus, there is no need to consider any transition dynamics over the trajectory and the resulting MDP has only one state.

In our scenario the reward brings information about the predictive ability of the trees of the Forest. Both the ground truth values and the predictions of the trees, which act like signals from the environment, contribute to the computation of the reward. Rewards are defined differently for binary classification and regression problems.

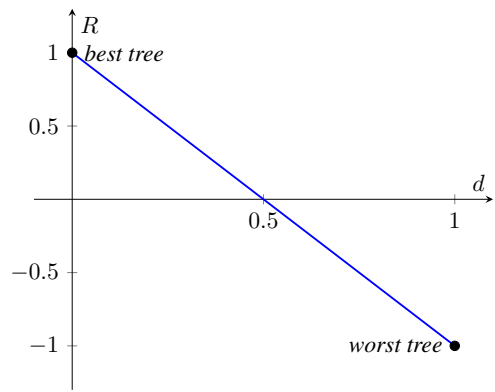
\begin{figure}
    \centering
    \resizebox{0.40\columnwidth}{!}{%
        \begin{tikzpicture}[
            declare function={
                func(\x) = (1 - 2*(\x));
            }]
            \begin{axis}[axis x line=middle, axis y line=middle,
                         ymin=-1.3, ymax=1.3, ytick={-1, -0.5, 0, 0.5, 1}, ylabel=$R$,
                         xmin=-0.1, xmax=1.1, xtick={0, 0.5, 1}, xlabel=$d$,
                         domain=0:1, samples=1000]
                \addplot[blue, thick] {func(x)};
                \filldraw[black] (110, 30) circle (2pt) node[inner sep=5pt, anchor=east]{\textit{worst tree}};
                \filldraw[black] (10, 230) circle (2pt) node[inner sep=5pt, anchor=west]{\textit{best tree}};
            \end{axis}
        \end{tikzpicture}
    }
    \caption{Reward used for regression problems. The two black dots specify the reward value with respect to the output of the best/worst tree of the forest for a particular example.}
    \label{fig:reward_drt}
\end{figure}

For classification problems the reward function is defined as:
\begin{equation} 
\label{eq:reward_ddt}
    r(a_{x}) = \begin{cases} 
                               +1 & h_{a_x}(x) = y \\
                               -1 & \text{otherwise},
                  \end{cases}
\end{equation}
where $y \in \lbrace -1, 1 \rbrace$ is the ground true value corresponding to the observed feature vector $x$ and $h_{a_{x}}(x) \in \lbrace -1, 1 \rbrace$ is the prediction provided by the tree selected through the action $a_x$.       
Concerning regression problems, the reward function depends on the squared prediction error $d(x,y) = (h_{a_x}(x) - y)^2$ as follows (see Fig. \ref{fig:reward_drt}):    
\begin{equation} 
\label{eq:reward_drt}
    r(a_x) = 1 - 2d
\end{equation}
where, for each example, $d$ is normalized in $[0, 1]$ respect to the squared errors of the best and worst tree in the forest. 

We motivate the use of a linear reward (w.r.t. the distance) rather than a step wise one to give greater importance to the well-aimed actions of the agent.

We choose to employ a DNN of parameters $\theta$ to approximate the policy (Deep RL).
In the training phase, we used the classical RL trial-error approach to update the policy parameters $\theta$ for every batch of examples (see Algorithm \ref{alg:policy update}).

The main problem in multi-armed bandits is the need for
balancing \textit{exploration} and \textit{exploitation}.
We want to avoid the agent to greedily select just those trees that seem to appear best, as they may in fact be suboptimal due to imprecision in the knowledge of the agent. Thus, during the training, we want the policy to explore the action space by choosing also seemingly not good trees to obtain more information about them.
For this reason we introduced a regularization term in our loss that gives importance also at the entropy of the distribution of the actions given the input. In this way, we are able to obtain a good \textit{exploration}/\textit{exploitation} trade-off, preventing the agent to rapidly converge to suboptimal tree recommendations.

\begin{algorithm}
    \caption{Policy Update} 
    \label{alg:policy update}
    \begin{algorithmic}
        \renewcommand{\algorithmicrequire}{\textbf{Input:}} \REQUIRE $\theta_k, \lbrace (x_i, y_i) \rbrace_{i=1}^N \; \text{(batch)}$
        \FOR{$i=1$ to $N$}
            \STATE Sample actions $a_{x_i} \sim \pi_{\theta_k}(a_{x_i}| x_i)$         \STATE Compute rewards $r(a_{x_i})$
        \ENDFOR
        \STATE Obtain $\mathcal{D} = \lbrace \big( x_i, r(a_{x_i}) \big) \rbrace_{i = 1}^N $
        \STATE Define $\mathcal{L}(\theta) = -\frac{1}{N}\sum_{i=1}^{N} \log\pi_\theta(a_{x_i}| x_i) r(a_{x_i})$ 
        \STATE Compute gradient $\nabla_{\theta} \mathcal{L}(\theta)$
        \STATE Update parameters $\theta_{k+1} = \theta_k - \alpha_k \nabla_\theta \mathcal{L}(\theta)$
        \RETURN $\theta_{k+1}$ 
    \end{algorithmic}
\end{algorithm}

Predictions on new data are obtained with the support of the trained RL system which dynamically selects trees based on the input data. In particular, for a given new instance $x_{\text{new}}$, the system detects a suitable tree predictor $h_{i^*} (\cdot)$ selecting the index $i^*$ as
\begin{equation*}
    i^* = \argmax_{i \in \mathcal{T}} \pi_{\theta}(a_{x_{\text{new}}}|x_{\text{new}}). 
\end{equation*}
Final predictions are computed following the path of decision rules from $h_{i^*}(\cdot)$.

\section{Experiments} \label{sec:numerical_experiments}


\subsection{Experimental setup}
We applied our method both for binary classification and regression problems. We used benchmark datasets from LIBSVM\footnote{\url{https://www.csie.ntu.edu.tw/~cjlin/libsvmtools/datasets/}} and UCI\footnote{\url{https://archive.ics.uci.edu/ml/datasets.php}} data repositories to assess our method. A description of datasets involved in our experiments is reported in Table \ref{tab:classification_dataset} and Table \ref{tab:regression_dataset}. 

\begin{table}[!ht]
    \centering
    \resizebox{0.6\columnwidth}{!}{%
        \begin{tabular}{ccc}
                              & \textbf{Number of Examples} & \textbf{Number of Features} \\ \hline\hline
            \textbf{PHISHING} & 11055                       & 68                          \\
            \textbf{BIODEG}   & 1055                        & 41                          \\
            \textbf{HEART}    & 270                         & 25                          \\
            \textbf{SPAM}     & 4601                        & 57                          \\
            \textbf{A2A}      & 2265                        & 119                        
        \end{tabular}}
    \caption{Overview of the datasets for binary classification}
    \label{tab:classification_dataset}
\end{table}

\begin{table}[!ht]
    \centering
    \resizebox{0.6\columnwidth}{!}{%
        \begin{tabular}{ccc}
                                & \textbf{Number of Examples} & \textbf{Number of Features} \\ \hline\hline
            \textbf{CALIFORNIA} & 20640                       & 8                           \\
            \textbf{FRIEDMAN}   & 5000                        & 20                          \\
            \textbf{YEARPRED}   & 463715                      & 90                          \\
            \textbf{ABALONE}    & 4177                        & 8                           \\
            \textbf{CPU}        & 8192                        & 12                         
        \end{tabular}}
    \caption{Overview of the datasets for regression}
    \label{tab:regression_dataset}
\end{table}

The aim of the experiments is to compare the predictive performances of CCT and CRT with the ones of CART, Random Forest (the one on which our method is built on) and a supervised baseline. We are mainly interested in a comparison with CART, which acts as our primary baseline, since it is an interpretable model and it is the standard algorithm used to train trees.     

Both for CART and Random Forest we set the maximum depth of trees at most equal to four.  
This choice is related to \textit{comprehensibility}: it is important that a human can rely on easily understandable explanations. This fact is dependent on psychological and social implications \cite{molnar2020interpretable}. Setting the maximum tree depth equal to four ensures also that the Miller's law \cite{Miller56Magic} holds. Furthermore, we grew up the Random Forest with fifty tree predictors: 
a forest with hundreds of trees would compromise the exploration phase of the agent, slowing down the convergence during the training of the policy.

\begin{table*}[ht!]
    \centering
    \resizebox{\columnwidth}{!}{%
        \begin{tabular}{cccc|cc}
                              & \textbf{Contextual Classification Tree}       & \textbf{Decision Tree} & \textbf{Supervised} &  \textbf{Random Forest} & \textbf{Tree Depth} \\ \hline\hline
            \textbf{PHISHING} & \textbf{0.9394 $\pm$ 0.001}          & 0.9209                  & 0.589 $\pm$ 0.009     & 0.9294           & 4                   \\
            \textbf{BIODEG}   & \textbf{0.8025 $\pm$ 0.016}          & 0.7867                   & 0.624 $\pm$ 0.012   & 0.8057            & 4                   \\
            \textbf{HEART}    & \textbf{0.7963 $\pm$ 0.030}          & 0.7778                   & 0.605 $\pm$ 0.038    & 0.8333           & 3                   \\
            \textbf{SPAM}     & \textbf{0.9233 $\pm$ 0.003}          & 0.8881                  & 0.706 $\pm$ 0.013    & 0.9077            & 4                   \\
            \textbf{A2A}      & \textbf{0.8293 $\pm$ 0.008}          & 0.8190                  & 0.718 $\pm$ 0.010    & 0.7748            & 3                  
        \end{tabular}}
    \caption{Accuracy and standard deviation for classification tasks. We report in bold the best results between CCT, CART (Decision Tree) and the Supervised approach. For the sake of completeness we also report here the results obtained by the Random Forest.}
    \label{tab:dynamic_decision_tree}
\end{table*}

\vspace{2em}
\begin{table*}[ht!]
    \centering
    \resizebox{\columnwidth}{!}{%
        \begin{tabular}{ccccc|cc}
                                & \textbf{Scale} & \textbf{Contextual Regression Tree}     & \textbf{Decision Tree} & \textbf{Supervised}  & \textbf{Random Forest} & \textbf{Tree Depth} \\ \hline\hline
            \textbf{CALIFORNIA} & $10^4$ & 7.456 $\pm$ 0.032         & 7.828                                & \textbf{7.420 $\pm$0.015} & 7.602   & 4                   \\
            \textbf{FRIEDMAN}   & $10^0$ & \textbf{3.014 $\pm$ 0.013}                 & 3.040                        &3.122 $\pm$0.008    & 2.718   & 4                   \\
            \textbf{YEARPRED}   & $10^1$ & \textbf{0.995 $\pm$ $<$ 0.001}           & 1.010                              &1.007$\pm$<0.001   & 1.001     & 4                   \\
            \textbf{ABALONE}   & $10^0$ & \textbf{2.379 $\pm$ 0.004}                   & 2.389                           &2.487 $\pm$ 0.045  & 2.307  & 4                   \\
            \textbf{CPU} & $10^0$        & \textbf{4.205 $\pm$ 0.026}          & 4.566                                   &4.281 $\pm$ 0.023  & 4.173   & 4
        \end{tabular}}
    \caption{Root Mean Squared Error and standard deviation for regression tasks. We report in bold the best results between CRT, CART (Regression Tree) and the Supervised approach. For the sake of completeness we also report here the results obtained by the Random Forest.}
    \label{tab:dynamic_regression_tree}
\end{table*}
\vspace{5em}

\begin{figure*}[ht!]
    \begin{center}
    \scalebox{1.0}{
    \begin{tabular}{ccc}
        \includegraphics[width=0.30\textwidth]{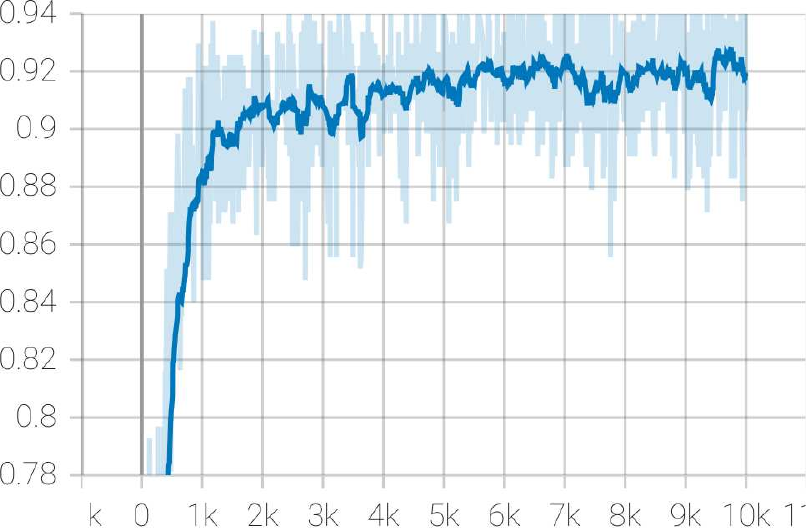} & 
       
        \includegraphics[width=0.30\textwidth]{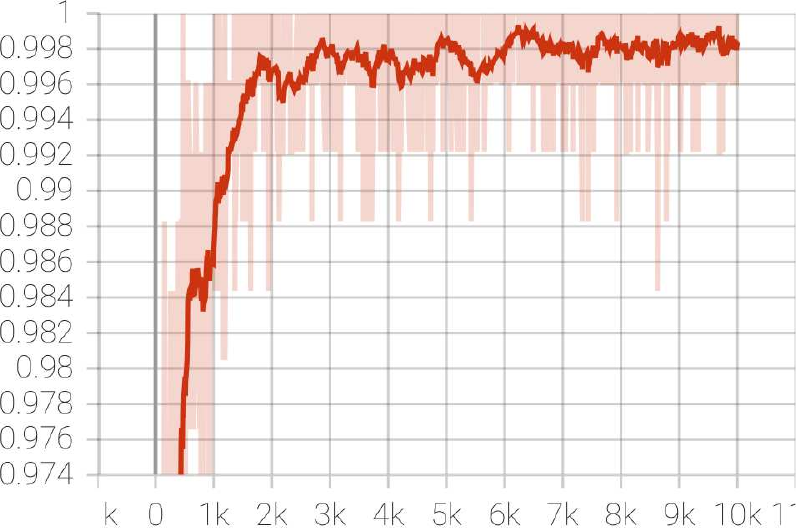} &  
       
        \includegraphics[width=0.30\textwidth]{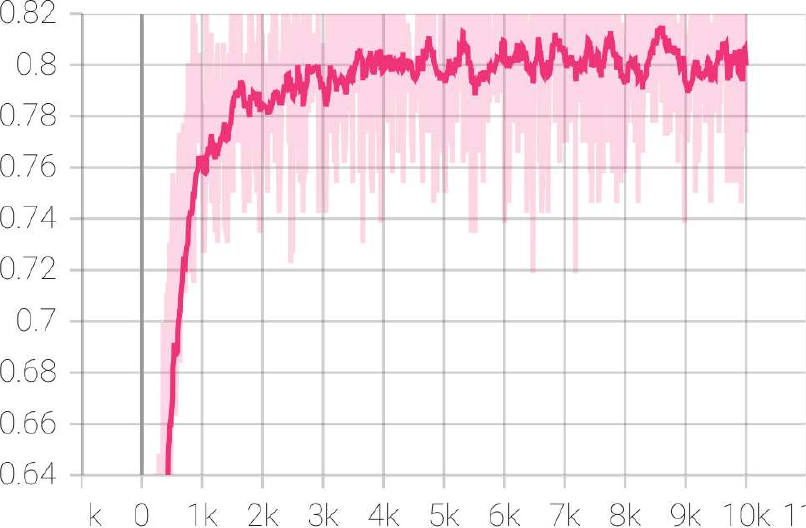} \\
        
        \includegraphics[width=0.30\textwidth]{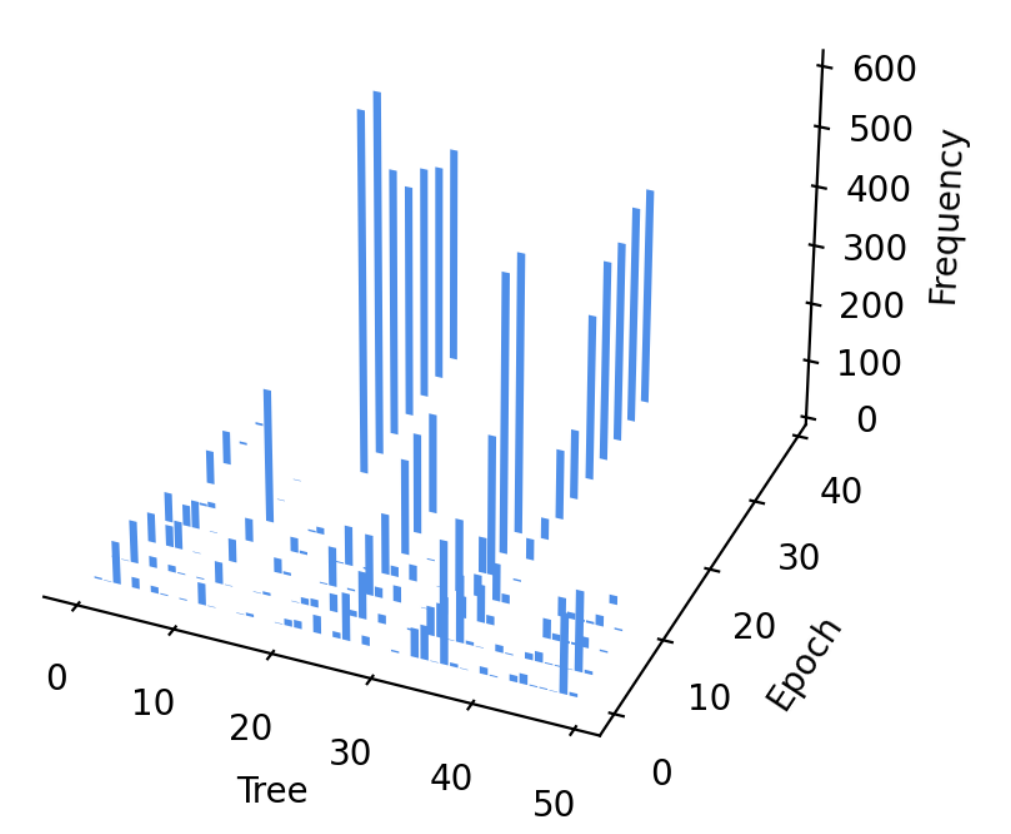} &
          
        \includegraphics[width=0.30\textwidth]{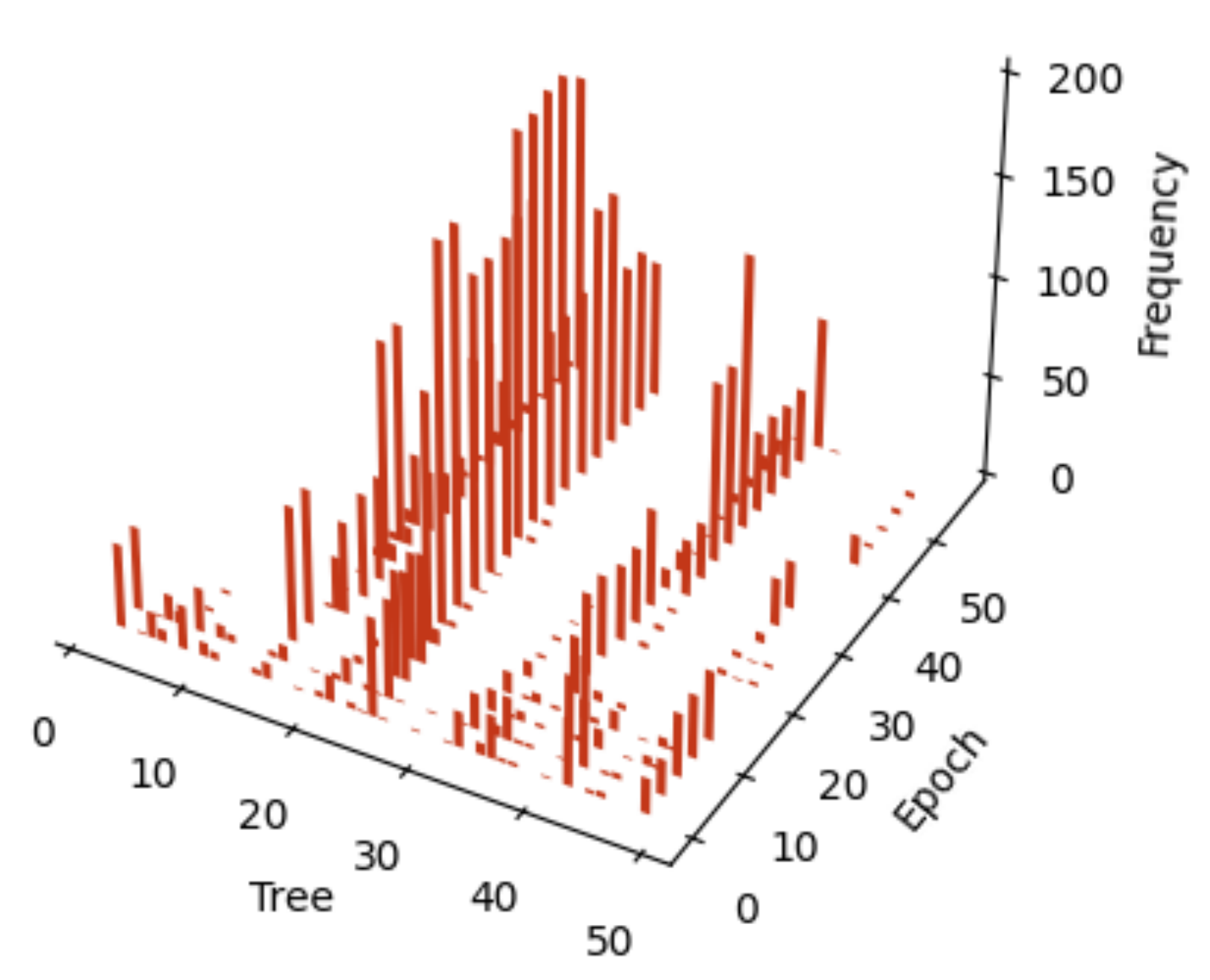} &
          
        \includegraphics[width=0.30\textwidth]{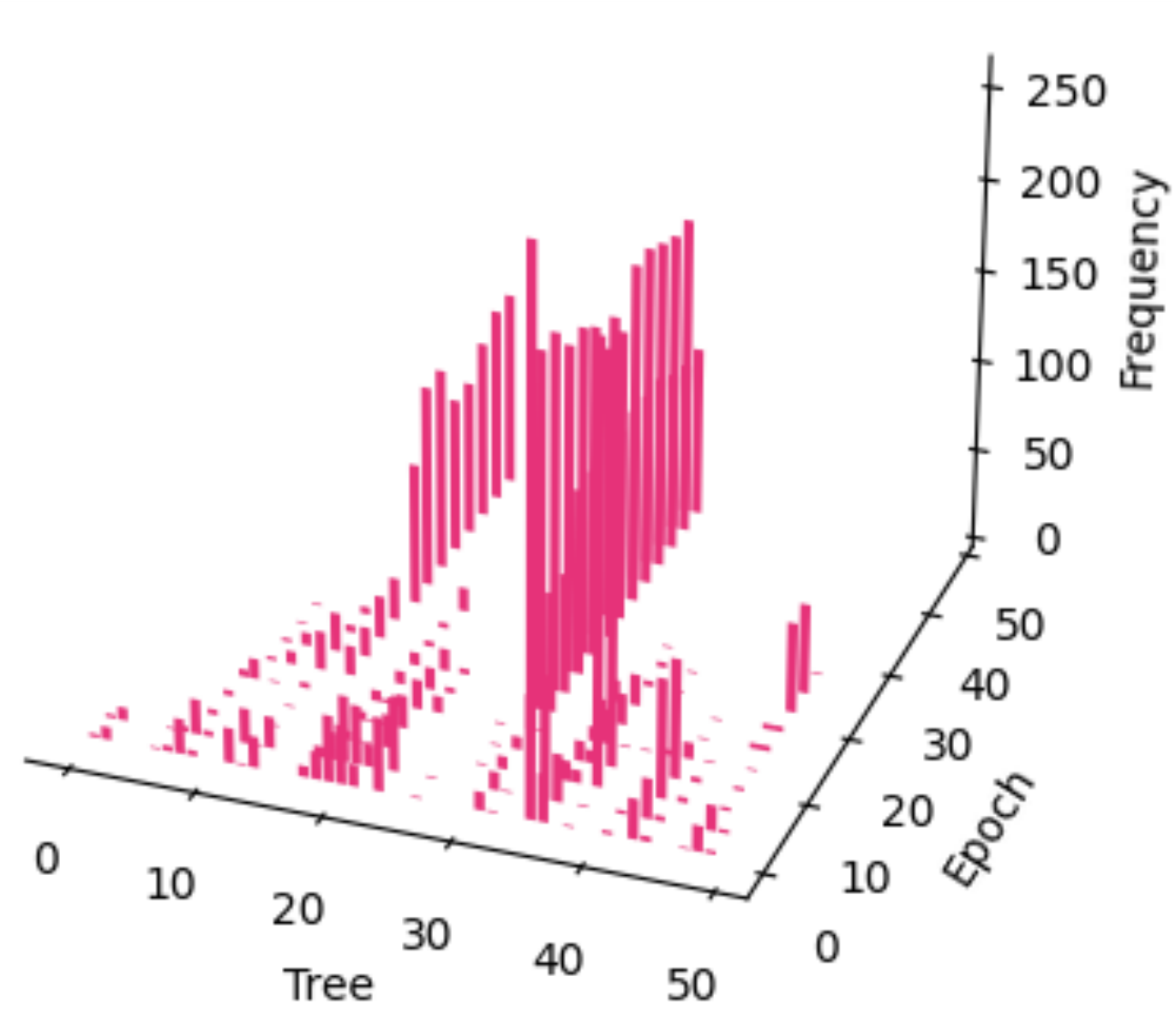} \\

        (a) & (b) & (c)\\
        
    \end{tabular}
    }
    \end{center}
    \caption{Reward (above) and Tree selection histograms (below) during training in classification. \\ (a) SPAM; (b) DIGITS; (c) A2A.}
    \label{fig:results}
\end{figure*}


The original whole dataset is split up in three parts (64 \% training set, 16 \% validation set, 20 \% test set) and the training phase consists of two steps. The first one involves the training of both CART and Random Forest, while the second is about the training of the neural network to approximate the policy.

We first used a grid search strategy, exploiting the validation set, to choose the best maximum depth for the baseline CART model. As anticipated above, in order to favour the interpretability, only maximum depths of 2, 3 and 4 have been considered as candidates. Subsequently, a Random Forest of 50 decision trees predictors, each grown up to the validated maximum depth, is trained on the training set. The other hyperparameters of both CART and Random Forest are set equal to their default values \cite{scikit-learn}. 

We used training data to learn the policy of tree recommendation. As network structure, we employed a 3-layers architecture with ReLU activations and a 0.2 Dropout layer \cite{srivastava2014dropout}. We would like to stress that the chosen network does not significantly impact the inference time of the whole model. An early stopping strategy \cite{montavon2012neural} is used to stop the training according to the predictive performances on the validation set. For both classification and regression problems we have used ADAM optimizer \cite{kingma2015adam} with an initial learning rate of $10^{-3}$ and no weight decay in combination with a cosine scheduler \cite{loshchilov2017SGDRSG} to slightly reduce the learning rate after each epoch.
Finally, we set the regularization term related to the entropy of the policy output distribution to $10^{-4}$, after validating it using a grid search strategy with respect to the values: $10^{-1}, 10^{-2}, 10^{-3}, 10^{-4}, 10^{-5}$.

Regarding the supervised baseline, we used the same network structure with a softmax activation in the last layer and the same hyperparameters. In classification context, we created the dataset labeling each sample with the label of the tree with the greatest probability among the set of the trees that lead to the correct prediction. In case there is no tree of the forest that correctly predicts the example, the label is randomly chosen according to a uniform distribution over the Random Forest trees. Similarly, in the case of regression, the chosen label is the one related to the tree in the forest that produces the smallest quadratic error respect to the target. We maintained the same dataset split configuration  and, also for this model, an early stopping strategy is used respect to the predictive performances on the validation set.

\subsection{Discussion}
The obtained predictive results are reported in Tables \ref{tab:dynamic_decision_tree}, \ref{tab:dynamic_regression_tree}. Results indicate that our method substantially outperforms CART and the supervised approach in both regression and classification tasks.

The supervised strategy has proved particularly unsuccessful in the classification and regression scenario due to the ill-defined mechanism of assignment of labels. We believe that a full supervised approach restricts the exploration of action space and adversely affects the ability to generalize on unseen data. 

Conversely, the more accurate predictive results, obtained by our method, are strongly linked to the capability of the agent to more thoroughly explore the action space. In this respect, the role of the entropy regularization is fundamental to encourage exploration. A weak regularization respect to the entropy of the action distribution inevitably would lead to a very greedy tree selection and therefore our method will tend to perform similar to the supervised baseline. Figure \ref{fig:results} highlights the trade-off between exploration and exploitation varying the training epochs. It is indeed possible to note two important things: \begin{inparaenum}[(i)]
	\item rewards do not rise considerably after 2000 epochs,
	\item after the first tens epochs, the RL agent learns to select the same two or three trees of the Random Forest.
\end{inparaenum}
This latter point evidences that the agent has got sufficient information about the reward obtained from the estimators and therefore the policy tends to converge choosing always a subset of trees that provides better performance. This means that only few estimators of the Random Forest contribute to the increase of the reward over the training epochs, basically excluding a large part of the trees. As a matter of fact, some works \cite{bernard2009selection} \cite{zhang2009search} \cite{latinne2001limiting} \cite{tripoliti2010dynamic}, although with different arguments, agree the existence of well performing sub-forests which constitute the core of the whole ensemble.

We also underline that the obtained results are particularly significant since good predictive accuracy is obtained without losing in terms of interpretability. In confirmation of this, we highlight that according to the recent methodology proposed in \cite{vidal2020born}, the generated \textit{optimal born-again tree} by construction reproduces the same decision function of the Random Forest and therefore equals but does not improve the predictive accuracy of the Random Forest. Moreover, the obtained predictor, despite having been constructed from an ensemble of only few (20 at most) and very shallow (depth of at most 3) trees, inevitably has a much greater depth than the starting ones, hence compromising the resulting interpretability.

\section{Conclusion} \label{sec:conclusion}
We proposed an innovative feature-based method which is able to recommend a single shallow tree in a Random Forest. We modelled this problem as a contextual multi-armed bandit and solved by RL methods. Since decision trees operate on input features to form their decision paths, our intuition is that an intelligent system able to appropriately recommend a single estimator of a grown Random Forest, can be advantageous in terms of predictive performance. A very important characteristic of our method is that an interpretable model is returned. More specifically, in order to compute a new prediction, differently from Random Forest, only one of its decision trees is in turn employed.

Our method outperforms CART and a supervised baseline in both regression and classification tasks and the obtained results are also comparable with the ones of Random Forest.

Interestingly, we observed that the system learned to choose the same two or three decision trees after the first tens of epochs. This fact is a good starting point for future research that may explore novel Random Forest pruning methods or novel re-weighting strategies.

\section*{Acknowledgements}
The authors would like to thank Prof. Andrew D. Bagdanov, Prof. Peter Frazier and Prof. Fabio Schoen for their personal advice, help and encouragement.

\bibliographystyle{unsrt}  
\bibliography{references}

\end{document}